\documentclass{article}

\usepackage{microtype}
\usepackage{graphicx}
\usepackage{subfigure}
\usepackage{booktabs} %

\usepackage{hyperref}

\usepackage{algpseudocode}

\usepackage{soul}

\usepackage{todonotes}

\usepackage{multirow}
\usepackage{graphicx}

\usepackage[accepted]{mlsys2024}

\mlsystitlerunning{SparAMX: Accelerating Compressed LLMs Token Generation on AMX-Powered CPUs}

\newif\ifcomment
\commenttrue

\ifcomment
\newcommand{\ahmed}[1]{\sethlcolor{cyan}\hl{[Ahmed: #1]}}
\newcommand{\moh}[1]{\sethlcolor{green}\hl{[Moh: #1]}}

\newcommand{\important}[1]{\sethlcolor{red}\hl{[IMP: #1]}}
\else
\newcommand{\ahmed}[1]{}
\newcommand{\moh}[1]{}
\newcommand{\important}[1]{}
\fi

\begin{document}

\twocolumn[
\mlsystitle{SparAMX: Accelerating Compressed LLMs Token Generation \\ on AMX-powered CPUs}

\mlsyssetsymbol{equal}{*}

\begin{mlsysauthorlist}
\mlsysauthor{Ahmed F. AbouElhamayed}{cornell}
\mlsysauthor{Jordan Dotzel}{cornell}
\mlsysauthor{Yash Akhauri}{cornell}
\mlsysauthor{Chi-Chih Chang}{cornell}
\mlsysauthor{Sameh Gobriel}{intel}
\mlsysauthor{J. Pablo Muñoz}{intel}
\mlsysauthor{Vui Seng Chua}{intel}
\mlsysauthor{Nilesh Jain}{intel}
\mlsysauthor{Mohamed S. Abdelfattah}{cornell}
\end{mlsysauthorlist}

\mlsysaffiliation{cornell}{Cornell University, New York, USA}
\mlsysaffiliation{intel}{Intel Labs, Oregon, USA}

\mlsyscorrespondingauthor{Ahmed F. AbouElhamayed}{afa55@cornell.edu}
\mlsyscorrespondingauthor{J. Pablo Muñoz}{pablo.munoz@intel.com}
\mlsyscorrespondingauthor{Mohamed S. Abdelfattah}{mohamed@cornell.edu}

\mlsyskeywords{Machine Learning, MLSys}

\vskip 0.1in

\begin{abstract}

Large language models have high compute, latency, and memory requirements. While specialized accelerators such as GPUs and TPUs typically run these workloads, CPUs are more widely available and consume less energy. Accelerating LLMs with CPUs enables broader AI access at a lower cost and power consumption. This acceleration potential for CPUs is especially relevant during the memory-bound decoding stage of LLM inference, which processes one token at a time and is becoming increasingly utilized with reasoning models. We utilize Advanced Matrix Extensions (AMX) support on the latest Intel CPUs together with unstructured sparsity to achieve a \textbf{1.42}$\times$ reduction in \textit{end-to-end} latency compared to the current PyTorch implementation by applying our technique in linear layers. We provide a set of open-source customized sparse kernels that can speed up any PyTorch model by automatically replacing all linear layers with our custom sparse implementation.
Furthermore, we demonstrate for the first time the use of unstructured sparsity in the attention computation achieving a \textbf{1.14}$\times$ speedup over the current systems without compromising accuracy. Code: \href{https://github.com/IntelLabs/Hardware-Aware-Automated-Machine-Learning/tree/main/SparAMX}{https://github.com/IntelLabs/Hardware-Aware-Automated-Machine-Learning/tree/main/SparAMX}

\end{abstract}
    \vspace{-8pt}
]

\printAffiliationsAndNotice{}  %

\section{Introduction}
\label{introduction}

The usage of large language models (LLMs) has grown exponentially over the past few years and is expected to continue its unprecedented growth. This has enabled many AI-driven applications, yet significant hardware and power resources are required, which have motivated recent attempts at model compression and acceleration. One such compression method is unstructured pruning, which removes some model parameters without structural constraints like contiguous nonzero values. Although such a method can achieve high sparsity while maintaining accuracy, achieving actual speedup on current hardware, such as GPUs and TPUs, is challenging.

\begin{figure} [t]
    \centering
    \includegraphics[width=1\linewidth]{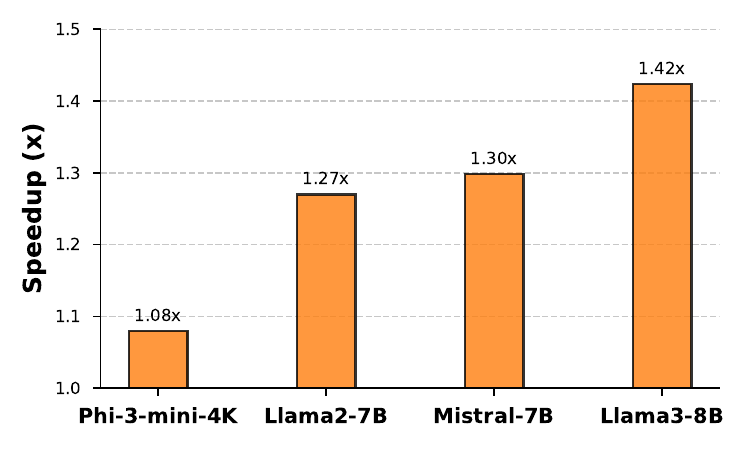}
    \vspace{-10pt}
    \caption{\textbf{SparAMX Performance:} SparAMX consistently leads to a speedup in decode latency over PyTorch. The improvement tends to be greater as the model size increases. Context length is set to 512.}
    \label{fig:multiple_models}
\end{figure}

This high-cost specialized hardware makes LLMs inaccessible to many people and limits their use. CPUs, on the other hand, are more ubiquitous and therefore can be used to accelerate LLM-driven applications for a wider audience. Newer CPUs, such as the Intel Sapphire Rapids, contain units that natively accelerate matrix multiplication at low cost and power. For example, AMX units present in the Sapphire Rapids chip enable direct acceleration of LLM workloads on the CPU. We explore combining the capabilities of this unit with unstructured sparsity to accelerate LLMs on the CPU. As shown in Figure~\ref{fig:multiple_models}, our sparse AMX kernel, SparAMX, leads to faster decoding times across common LLMs compared to the stock PyTorch.

In this work, we start with background material on sparsity and the hardware features used in our kernel, explore the current literature, and then introduce our detailed optimizations and design of kernels. We then demonstrate better efficiency over stock PyTorch that uses Intel's optimized libraries \cite{onednn} and other proprietary commercial solutions like DeepSparse \cite{deepsparse} by optimizing the dominant matrix multiplication operations within the model's linear layers. Finally, we apply unstructured pruning to the KV cache for the first time and propose a kernel that makes use of the sparsity to optimize the attention operation as well to increase model performance.

Our summarized contributions are as follows:

\begin{itemize}
    \item An end-to-end system that uses unstructured sparsity to improve latency by up to \textbf{1.42$\times$} over the stock PyTorch on CPU.
    \item An INT8 CPU kernel that uses unstructured sparsity and AMX to achieve up to \textbf{1.46$\times$} better performance than current proprietary kernels~\cite{deepsparse} for quantized models.
    \item A novel study of unstructured sparsity in the KV cache achieving \textbf{1.14$\times$} speedup on 16K context with minimal accuracy loss.

\end{itemize}

\section{Background}

\begin{figure} [t]
    \centering
    \includegraphics[width=1\linewidth]{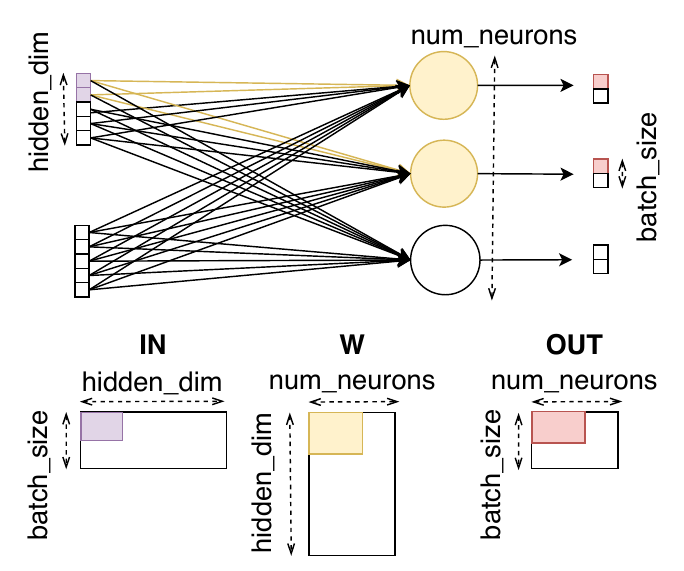}
    \vspace{-15pt}
    \caption{\textbf{Linear GEMM Mapping:} A linear layer can be computed with matrix multiplication. In this example hidden\_dim = 4, num\_neurons = 3, and batch\_size = 2.}
    \label{fig:linear_as_matmul}
\end{figure}

\subsection{LLM Inference}
\label{sec:llm_background}

LLM inference consists of two distinct stages with varying compute and memory requirements. During the first stage, known as \textit{prefill}, all input tokens are processed at the same time, allowing significant memory reuse and making the operation \textbf{compute-bound}. However, this stage occurs only once per prompt. The needed results (keys and values of each token in each attention head in each layer) are stored in the KV cache for subsequent use during the second stage.

The second stage, known as \textit{decode}, performs auto-regressive sampling one token at a time. This token-by-token processing reduces memory reuse and compute intensity, making the decode stage \textbf{memory-bound} and bottlenecked by weight loading from memory in short contexts.
This opens an opportunity to leverage unstructured sparsity, which enables higher sparsity levels and more efficient compression to minimize memory transfer. Weights can be stored in a compressed format, retrieved with reduced memory overhead, and decompressed on the compute unit before performing the dense matrix multiplication. 

While this approach introduces some additional computation, the process is memory-bound; thus, reducing memory transfer, even at the cost of extra computation, results in an overall speedup. We propose a generic system that applies this optimization and integrates seamlessly with PyTorch. Our approach demonstrates a significant speedup compared to the current PyTorch implementation.

\subsection{DNN Sparsity}
\label{sec:background_sparsity}

In \textbf{Deep Neural Networks (DNNs)}, some weights contribute more significantly to the model's performance than others. Prior research has extensively explored methods for identifying weight importance~\cite{liu2023deja,akhauri2024shadowllm}. Introducing sparsity into the weights via pruning leverages this insight to reduce the size of the DNN while improving efficiency. Sparsity reduces memory transfer costs by avoiding the loading of insignificant weights and lowers compute requirements by skipping operations involving those weights.

There are two types of sparsity: \textbf{structured} and \textbf{unstructured}. 

\begin{itemize}
    \item \textbf{Structured sparsity} involves pruning weights in full blocks, such as a full row (mapping to a full neuron in Figure \ref{fig:linear_as_matmul}) or a full tile. This pattern allows for simpler acceleration since entire sets of weights can be skipped during loading and computation.
    \item \textbf{Unstructured sparsity} imposes no constraints on the patterns of pruned weights. This flexibility enables higher sparsity levels, but achieving actual speedup is more challenging because the irregular distribution of zeros requires full computation and weight loading unless some special handling is used.
\end{itemize}

Structured sparsity typically results in lower sparsity levels because important weights often co-exist within the same structure as prunable weights, preventing their independent removal. In contrast, unstructured sparsity achieves greater sparsity due to its flexibility, making it a better approach for model compression in scenarios where lower data transfer is desired and where hardware can handle its computational challenges.

\begin{figure} [t]
    \centering
    \includegraphics[width=1\linewidth]{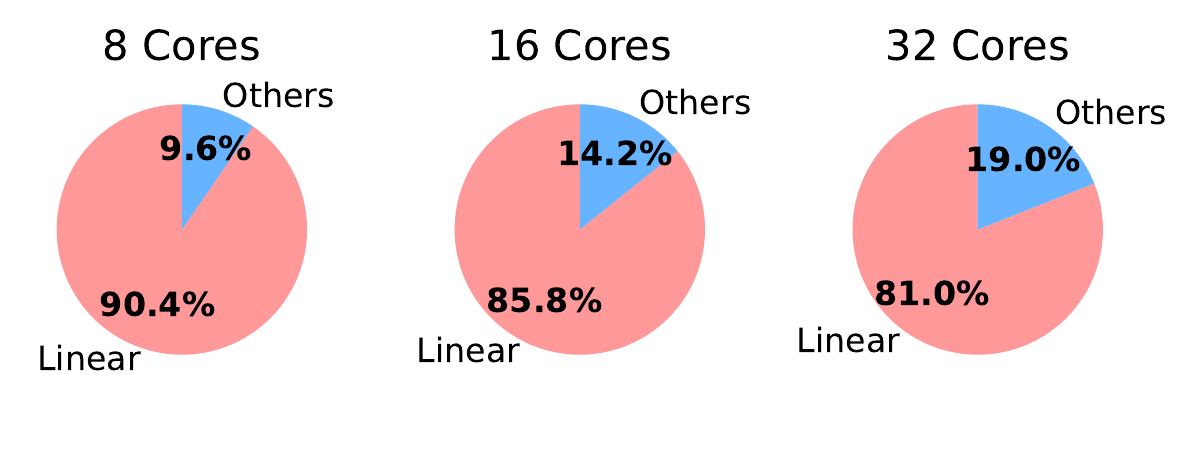}
    \vspace{-20pt}
    \caption{\textbf{Inference Breakdown:} Linear layers dominate the latency during LLM inference. Results profiled from Llama 3 8B running on Intel(R) Xeon(R) Gold 6430L CPU with 512 context length. }
    \label{fig:layer_breakdown}
    \vspace{-16pt}
\end{figure}

\subsection{GEMM Mapping}
\label{sec:gemm_mapping}

During both inference stages, the most compute-intensive operations in LLMs are mapped to matrix multiplication, also known as General Matrix Multiply (GEMM). These operations dominate the computational workload in both the core linear layers and attention mechanisms. Figure~\ref{fig:layer_breakdown} shows that linear layers dominate the latency, especially at small contexts.

Linear layers in LLMs can be represented as matrix multiplications, as illustrated in Figure~\ref{fig:linear_as_matmul}. In the example, the weight matrix (\textbf{W}) contains the weights, where each column corresponds to the weights of a single neuron. Each neuron uses its unique set of weights to process inputs. The input matrix (\textbf{IN}) contains the input values, with each row representing an individual input. Each input undergoes identical processing, being multiplied by the same set of weights.
The result of this computation is the output matrix (\textbf{OUT}). Each column in \textbf{OUT} corresponds to a neuron, with its rows containing the computed outputs for each input. The number of rows is the same as the number of inputs as an output row is computed for each of the input rows.

\subsection{Advanced ISA Extensions}

Our kernel is designed to utilize two specialized instruction sets: \textbf{AVX (Advanced Vector Extensions)} and \textbf{AMX (Advanced Matrix Extensions)}.

AVX is a set of SIMD (Single Instruction, Multiple Data) instructions extending the x86 architecture, enabling parallel operations on data vectors. It uses special AVX registers to store arguments and outputs for these operations. This work focuses on AVX-512, which operates on 512-bit registers.

For example, the instruction \textit{mm512\_loadu\_si512} can load 512 bits of data from main memory into an AVX register. To perform a dot product, you can load elements from two vectors into AVX registers and use the \textit{mm512\_dpbf16\_ps} instruction. This operation multiplies 32 pairs of 16-bit elements of the two registers, adds the results of each two consecutive elements together to form 16 32-bit elements which are accumulated in a third register.

Building on the success of AVX, Intel introduced AMX in their Sapphire Rapids processor series. Unlike AVX, which focuses on vectorized operations, AMX provides specialized hardware acceleration for matrix multiplication.

\begin{figure} [t]
    \centering
    \includegraphics[width=1\linewidth]{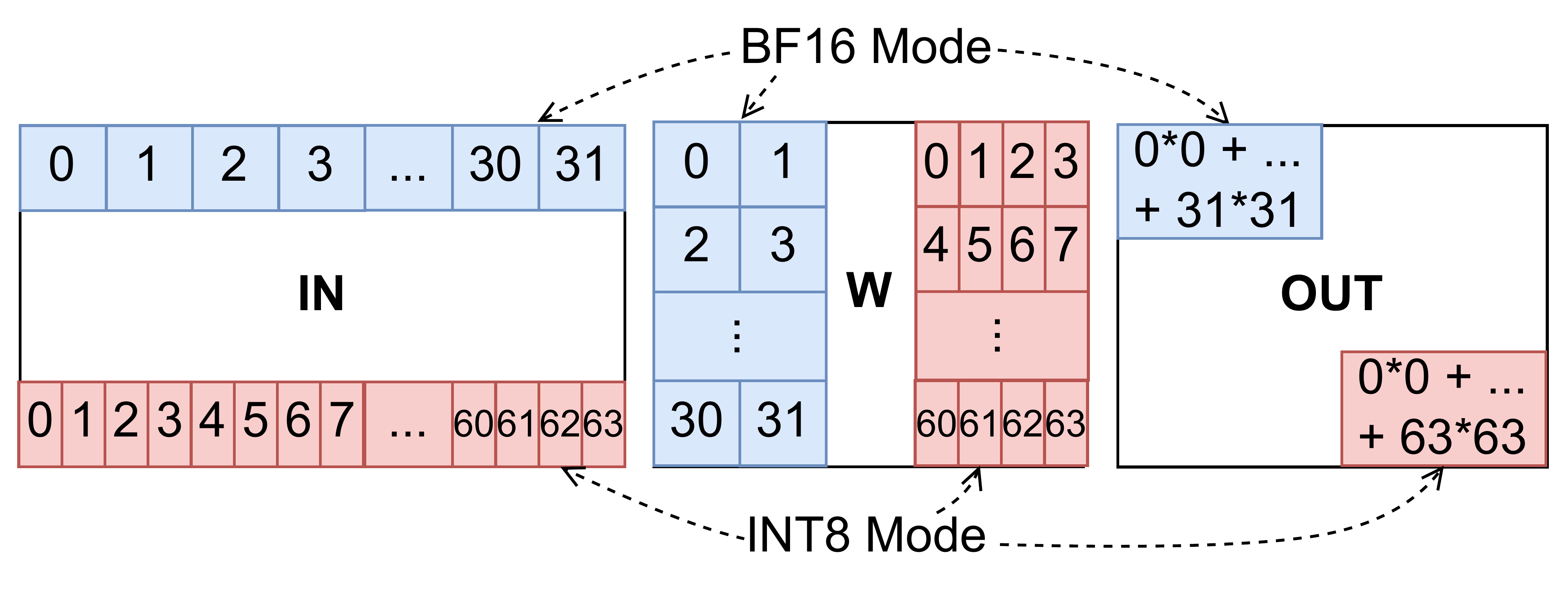}
    \vspace{-10pt}
    \caption{\textbf{AMX Matrix Multiplication:} AMX tiles support BF16 (blue) and INT8 (red) values. First two tiles contain the input matrix tiles, which are then multiplied and accumulated to the third tile.}
    \label{fig:amx_tile}
\end{figure}

AMX introduces a set of tiled registers, which are two-dimensional and significantly larger than AVX registers (as shown in Figure~\ref{fig:amx_tile}). Each tile consists of up to 16 rows, with each row containing up to 512 bits, and there are eight tile registers in each AMX unit. These tiles support operations to zero all elements, load data from memory, store data into memory, and perform matrix multiplication.

Matrix multiplication in AMX involves multiplying two tiles and accumulating the result in a third tile. These operations support two data formats:
\begin{itemize}
    \item \textbf{BF16 (Bfloat16):} Each row can store up to 512 ÷ 16 = 32 elements. Results are stored in \textbf{FP32}.
    \item \textbf{INT8 (8-bit integers):} Each row can store up to 512 ÷ 8 = 64 elements. Results are stored in \textbf{INT32}.
\end{itemize}
    \vspace{-8pt}

During matrix multiplication, each word from the first operand tile is multiplied by the corresponding word from the second operand tile, requiring special arrangement of items in the matrices to match the ordering shown in Figure~\ref{fig:amx_tile}.

\section{Related Work}

\textbf{Quantization} Weight quantization in LLMs has been extensively studied, achieving high model quality with as few as three or four bits~\cite{lin2024awq,frantar2022gptq,dettmers2022gpt3, dotzel2024students}. Quantization of both weights and activations has also advanced significantly, enabling four-bit computation while maintaining accuracy~\cite{xiao2023smoothquant, liu-etal-2023-llm}. 

Given that trend, support for low-bit datatypes and operations has been added in recent CPUs and GPUs. For example, Intel Sapphire Rapids CPUs support INT8 operations, and NVIDIA's H100 Tensor Cores support INT8 and INT4 operations. To harness these features, specialized kernels have been developed, demonstrating up to a 3$\times$ speedup over full-precision models~\cite{lin2024awq}, with further optimizations achieving an additional 1.9$\times$ speedup~\cite{kim2024quick}.

Some kernels, such as ~\cite{shen2023efficient} efficiently run 4-bit models on multiple Intel platforms, including AMX, but do not incorporate sparsity.

\begin{figure*} [t]
    \centering
    \includegraphics[width=1\linewidth]{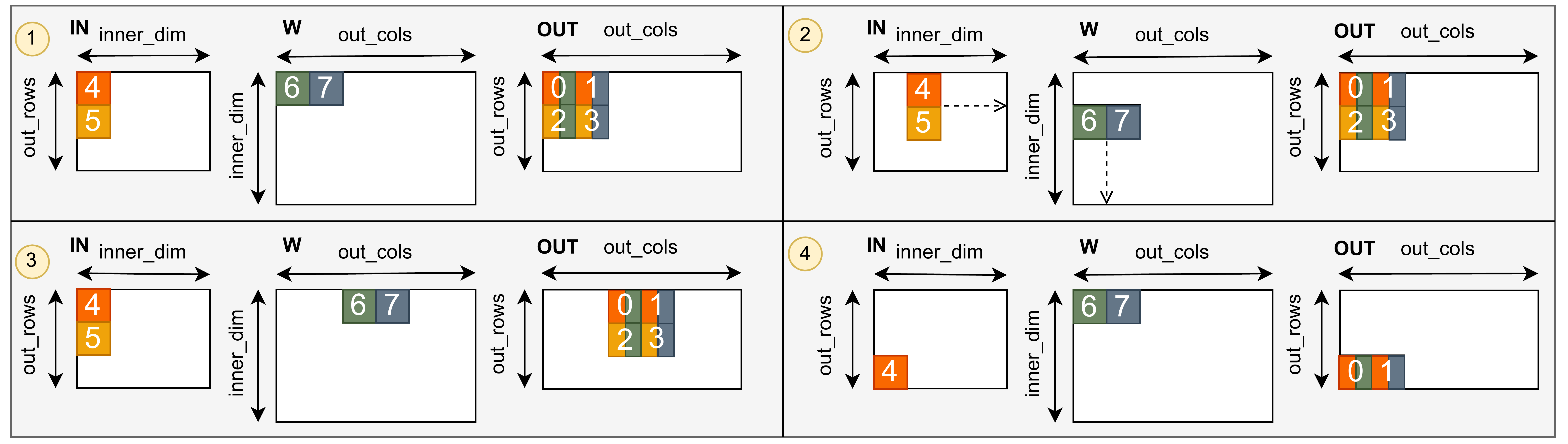}
    \vspace{-10pt}
    \caption{\textbf{AMX Dense Kernel: }(1) Tiles 0, 1, 2, and 3 act as accumulators for the results computed by multiplying (4x6), (4x7), (5x6), and (5x7) respectively. (2) Tiles 4 and 5 are utilized to load all columns of the input rows (denoted as out\_rows) while tiles 6 and 7 are utilized to load all rows of the weight columns. (3) After the loop over the inner dimension ends, results are stored in memory and a new set of result tiles is initialized and computed in the same way. (4) Some boundary conditions might occur near the end of the matrix.}
    \label{fig:dense_kernel}
\end{figure*}

\textbf{Sparsity} Like quantization, sparsity can be used to accelerate LLM workloads, as introduced in Section~\ref{sec:background_sparsity}. Many studies focus on static structured sparsity~\cite{ma2023llm, dery2024everybody}, which removes entire components, often yielding strong speedups at the cost of accuracy. For instance, LLM-Pruner~\cite{ma2023llm} performs one-shot structured pruning using gradient information and reaches only 20\% sparsity with an extra perplexity of 3.6 points on WikiText-2~\cite{merity2016pointer}. 
Other works leverage dynamic structured sparsity to optimize the runtime of LLMs on the GPU~\cite{liu2023deja, akhauri2024shadowllm}.

For unstructured sparsity, techniques such as SparseGPT~\cite{frantar2023sparsegpt} and Wanda~\cite{sunsimple} achieve up to 50\% pruning. However, these methods often incur accuracy losses—for example, 3.47\% for SparseGPT on Llama 2 7B. Wanda’s semi-structured variant suffers even greater perplexity degradation (+2.69) compared to its unstructured counterpart (+1.01) ~\cite{zhu2023survey}. Some techniques like Shears~\cite{munoz2024shears} can recover some of the lost accuracy with additional fine-tuning.
As wall-time acceleration of unstructured sparsity can be difficult, Flash-LLM~\cite{xia2023flash} proposes a GPU kernel that makes use of unstructured sparsity to minimize memory transfer requirements by cleverly using the general purpose GPU streaming multi-processors and tensor cores, while Marlin~\cite{frantar2024marlin} develops a GPU kernel that makes use of both quantization and semi-structured sparsity to improve performance.

While most prior works focus on GPUs, recent research has explored unstructured sparsity on CPUs. SparseDNN~\cite{wang2021sparsednn} uses static code generation to process only non-zero elements efficiently. DeepSparse Engine~\cite{deepsparse} applies unstructured sparsity with additional optimizations to accelerate LLMs, though it does not leverage AMX units and is closed-source. Our work integrates unstructured sparsity with AMX on Sapphire Rapids CPUs to improve LLM performance. Furthermore, we extend the unstructured sparsity to compress the KV cache and show 1.14$\times$ speedup using our kernel in the attention mechanism.
Some recent work showed that utilizing a CPU with AMX can help optimize the performance of a CPU-GPU system~\cite{10538369}. Our work aims to make running operations utilizing AMX even faster.

\section{Kernel Design}
\label{sec:kernel_design}

We design our kernels as PyTorch C++ extensions, accompanied by Python classes, enabling seamless replacement of layers in arbitrary PyTorch models. These kernels are general-purpose and do not assume or optimize for a specific sparsity pattern. Consequently, the achieved speedup depends on the sparsity percentage of the model.

We begin by introducing a dense kernel that performs standard GEMM operations in BF16 using AMX. We then extend this kernel to incorporate unstructured sparsity, adapting it to improve performance under sparse conditions. To evaluate performance gains of AMX, we also implement the sparse kernel using AVX and use it for comparisons. Finally, we present the quantized INT8 kernels, designed to leverage low-bit computation for further optimization.

\subsection{Dense Kernel}
\label{sec:desnse_kernel}

We begin by developing a kernel for linear layer computation using AMX, without incorporating any custom optimizations. This kernel, referred to as the \textbf{
dense kernel}, assumes a fully dense model with no sparsity-related modifications, as illustrated in Figure~\ref{fig:dense_kernel}. The AMX unit in each core can hold up to eight distinct tiles simultaneously. In our design:
\begin{itemize}
    \item Tiles 0-3 are utilized to store intermediate results, which remain in the AMX unit during iteration over the inner dimension.
    \item During each iteration, we load two input tiles (Tiles 4 and 5) and two weight tiles (Tiles 6 and 7), compute the matrix multiplication, and accumulate the results in the four result tiles.
    \item Upon completing the inner dimension loop, the four result tiles are saved to memory, and the next set of four result tiles is initialized.
\end{itemize}

By leveraging all eight tiles, instead of the naive approach of using one tile for result and two for operands, we achieve a compute-to-load ratio of 1:1, improving significantly over the 1:2 ratio that results from computing one tile at a time after loading two.

Next, we parallelize the operations in the kernel. Since each input row (representing a different input token) and output column (representing a neuron) is independent, parallelization can be applied over $out\_rows$ and $out\_cols$. In the decoding stage, a single batch contains only one output row. On the other hand, $out\_cols$ is input-independent and layer-dependent, making it the preferred dimension for parallelization. For smaller models, where $out\_cols$ is less than 32$\times$ number of available threads (since two tiles are processed at a time, each with 16 columns), parallelizing over $out\_rows$ as well can be beneficial. Parallelization over the inner dimension is another option, but it requires combining partial results from multiple result tiles, adding computational and memory overheads. This approach is only effective when both $out\_cols$ and $out\_rows$ are small and do not fully utilize all available threads.

\subsection{Sparse Format}

\begin{table}
    \centering
        \caption{\textbf{Compute Analysis:} Percentage of pipeline slots that were memory bound (waiting on any memory access, including caches) and DRAM Bound (missed the cache and waiting for main memory access). Evaluated on 32 consecutive linear layers with 4192 inputs and 14336 outputs to mimic Llama 3 8B up\_proj layer.}
    \begin{tabular}{ccc}
        \toprule
         Kernel& Memory Bound (\%) & DRAM Bound (\%) \\
        \midrule
         Dense& 100& 87.5 \\
         Sparse& 21.1& 5.7 \\
        \bottomrule
    \end{tabular}
    \label{tab:memory_vs_compute_dense_vs_sparse}
\end{table}

While the dense kernel leverages AMX efficiently, the analysis in Table~\ref{tab:memory_vs_compute_dense_vs_sparse} highlights further opportunities for performance optimization. Using the VTune Profiler to profile a Llama 3 model layer repeated 32 times, we observe that most pipeline slots are memory-bound, with nearly 50\% of the time spent waiting on slow DRAM access. To address this bottleneck, we reduce memory transfer by storing weights in a compressed format and decompressing each tile only when needed for computation.

Typically, weights, whether zero or non-zero, use the same number of bits. To save memory and bandwidth, we employ a compressed format where zero weights are represented with a single bit, while non-zero weights require an additional bit. Before computation, these values are converted back to their original format. Although this approach incurs some computational overhead during format conversion, it significantly reduces memory transfer, making it highly effective for the memory-bound decode phase.

\begin{figure} [t]
    \centering
    \includegraphics[width=1\linewidth]{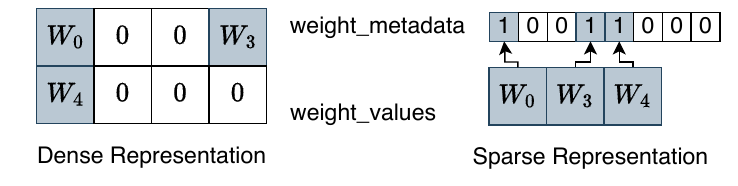}
    \vspace{-10pt}
    \caption{\textbf{Sparse Format:} Weights are stored with a sparse format that stores only the non-zero weights and an associated bitmap to indicate corresponding weight indices.}
    \label{fig:sparse_representation}
\end{figure}

This compressed weight format is illustrated in Figure~\ref{fig:sparse_representation}. The representation is divided into two components:
\begin{itemize}
    \item $weight\_metadata$: A bitmap where 1 indicates non-zero weights and 0 represents zero weights.
    \item $weight\_values$: A list containing all non-zero weights in the order required for processing.
\end{itemize}

We select this method for its simplicity and because modern hardware supports efficient extraction from this format.

 \begin{figure*} [t]
    \centering
    \includegraphics[width=1\linewidth]{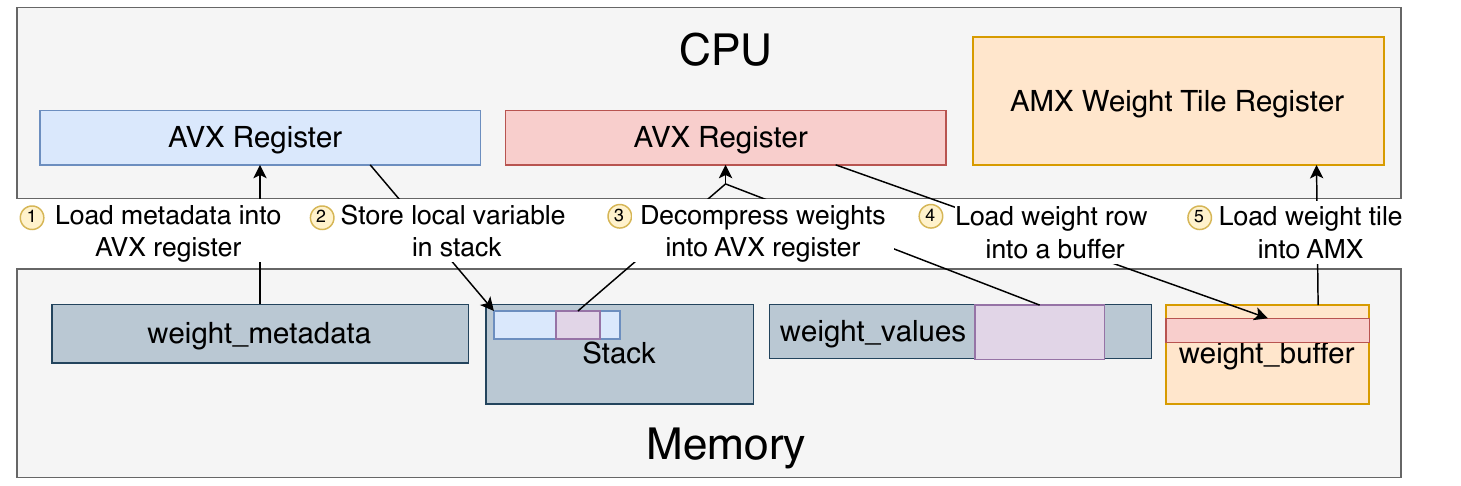}
    \vspace{-13pt}
    \caption{\textbf{Decompression Process:} To construct a single weight tile in case of BF16 computation, (1) 16 32-bit metadata elements are fetched into an AVX register. (2) The values in the local AVX register are stored in the stack and (3) accessed individually to guide the decompression into another AVX register. (4) The data in the register is stored in the $weight\_buffer$. (5) After all 16 weight rows are stored in the buffer, the data in the buffer is loaded into the AMX tile. See Appendix~\ref{app:sparse_weights_decompression_process} for algorithm version.}
    \label{fig:decompression_process}
\end{figure*}

\begin{figure*} [t]
    \centering
    \includegraphics[width=1\linewidth]{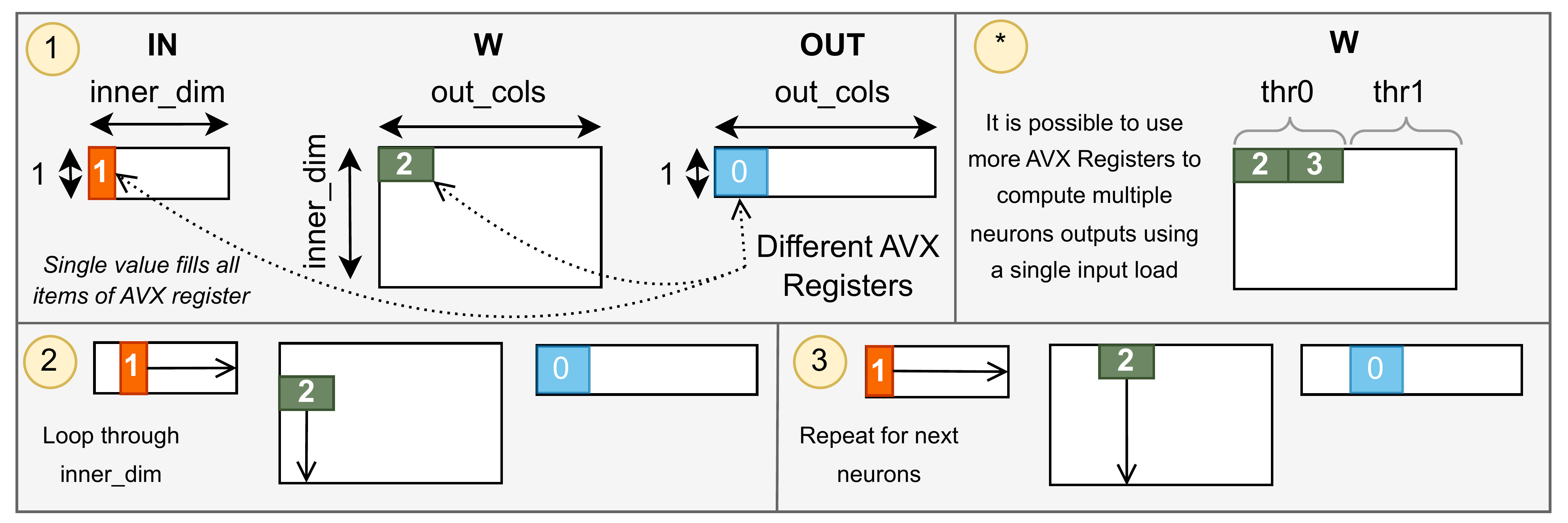}
    \vspace{-10pt}
 \caption{\textbf{AVX Matrix Multiplication: }Weights for different neurons are loaded into AVX register \textit{1} and a single element of the input is fetched and used to fill all elements of AVX register \textit{2}. AVX register \textit{0} is initialized to 0 to store the result, AVX registers \textit{1} and \textit{2} are multiplied and the results are added to the elements in AVX register \textit{0}. The process is repeated along the hidden dimension till all weights of the considered neurons are multiplied by the corresponding parts of the input. Another set of neurons are then processed.}
\label{fig:avx_implementation}
\end{figure*}

\subsection{AMX Sparse Kernel}
\label{sec:amx_sparse_kernel}

\begin{figure} [t]
    \centering
    \includegraphics[width=1\linewidth]{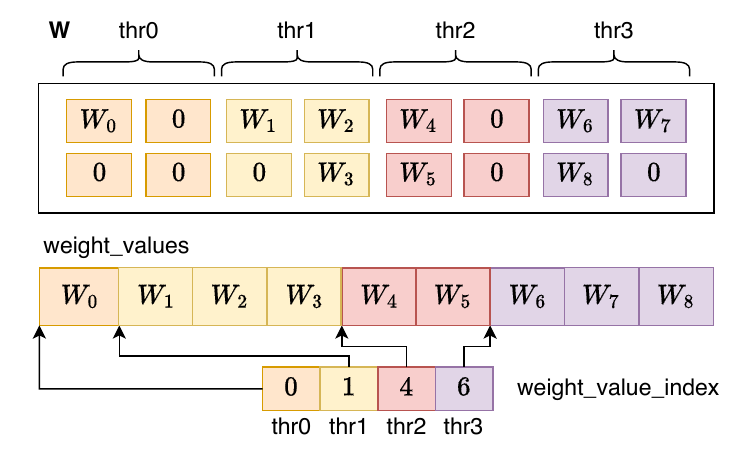}
    \vspace{-20pt}
    \caption{\textbf{Sparse Kernel Parallelization:} Parallelization over threads requires each thread to know where to begin processing the weights. The index list $weight\_value\_index$ stores this information and is computed offline allowing efficient parallelization at run time.}
    \vspace{-10pt}
    \label{fig:parallelizing_sparse}
\end{figure}

Parallelizing over weights compressed in this format is challenging due to its unstructured nature. In a multi-threaded program, the access points for each thread within $weight\_values$ are not predetermined. To address this, we introduce a precomputed index list, $weight\_value\_index$, generated during model initialization. This list specifies the starting position for each thread within $weight\_values$, as illustrated in Figure~\ref{fig:parallelizing_sparse}. Consequently, the number of threads must remain fixed during initialization, introducing a one-time compute overhead during model loading. However, this overhead occurs only once, and the runtime memory overhead is minimal, as it requires storing just one index per thread so it is practical for real-world application.

The kernel computation follows the dense kernel in Section~\ref{sec:desnse_kernel}, with the primary difference being that weights are not directly loaded into AMX registers because they are stored in a compressed format. Instead, the $weight_metadata$ items corresponding to the current weight tile are first loaded into an AVX register using the $vmovdqu32$ operation as shown in (Figure~\ref{fig:decompression_process}), which loads 512 bits of data as 16 groups of 32-bit elements. Each 32-bit element, along with a pointer to the index of the first non-consumed $weight\_values$ item ($weight\_value\_index$), serves as input to the $vpexpandw$ operation. This operation expands the bitmask into individual weights, placing zeros in positions marked by 0s in $weight_metadata$ and loading the corresponding weights for positions marked by 1s.

With a tile size of 16 rows by 32 columns, each 32-bit element expands to fill a row, and $vpexpandw$ is applied 16 times to populate the entire weight tile. The result of each $vpexpandw$ is stored temporarily in memory because direct transfers from AVX to AMX registers are currently unsupported. Instead, the complete tile is stored in memory and then loaded into AMX registers via an AMX load operation. Although this process involves memory access, frequent reuse of this memory region likely ensures it remains in the cache, mitigating expensive memory operations.

To update the $weight\_value\_index$ pointer, we utilize the $popcount$ operation. For efficient instruction-level parallelism, each of the 16 loop iterations is kept independent. The offset for each weight row is computed using $vpopcntd$, which calculates a $popcount$ on each of the 16 32-bit elements and stores the results in a new AVX register. Finally, a parallel prefix sum operation, as described in Algorithm~\ref{alg:prefix_sum}, computes the final offset needed for each tile row.

\begin{algorithm}[t]
\caption{Parallel Prefix Sum with AVX-512 Intrinsics}
\label{alg:prefix_sum}
\begin{algorithmic}
\State \textbf{Input:} AVX Register $v = [v_0, v_1, \dots, v_{15}]$ of 16 32-bit integers
\State \textbf{Output:} AVX Register $s = [s_0, s_1, \dots, s_{15}]$ where $s_i = \sum_{j=0}^{i} v_j$
\State
\State $s \gets v + \texttt{Shift}(v, 1)$ \Comment{Shift right 1 element and add\phantom{s}}
\State $s \gets s + \texttt{Shift}(s, 2)$ \Comment{Shift right 2 elements and add}
\State $s \gets s + \texttt{Shift}(s, 4)$ \Comment{Shift right 4 elements and add}
\State $s \gets s + \texttt{Shift}(s, 8)$ \Comment{Shift right 8 elements and add}
\State \Return $s$
\end{algorithmic}
\end{algorithm}

\subsection{AVX Sparse Kernel}

During the decode phase with batch size = 1, only one row of the 16-row AMX tile used for input is utilized, leading to significant inefficiency. To address this, we implement the compression technique using only AVX instructions, enabling a performance comparison between AVX and AMX while ensuring compatibility with older CPUs lacking AMX support. The flow is illustrated in Figure~\ref{fig:avx_implementation}. In this setup:
\vspace{-8pt}
\begin{itemize}
    \item An AVX register holds the weights for multiple neurons in the inner dimension.
    \item Another AVX register holds the corresponding input value repeated across the register.
    \item A third AVX register accumulates results for multiple neurons.
\end{itemize}

Further implementation details are provided in Appendix~\ref{app:avx_kernel_optimization}.

\subsection{INT8 Kernels}

AMX supports both BF16 and INT8 operations. We develop a kernel for INT8 matrix multiplication and adapt our framework to quantize weights and activations for compatibility. The flow of the INT8 kernel mirrors that of the AMX dense and sparse kernels, with adjustments for 8-bit elements instead of 16-bit. 

Each weight tile now contains $16 \times 64 = 1024$ weights, and their metadata is fetched into two AVX registers, each covering eight rows of the tile. Additionally, the weight ordering during preprocessing is modified to divide each weight column into four segments rather than two.

\section{Results}

\begin{figure} [t]
    \centering
    \includegraphics[width=.99\linewidth]{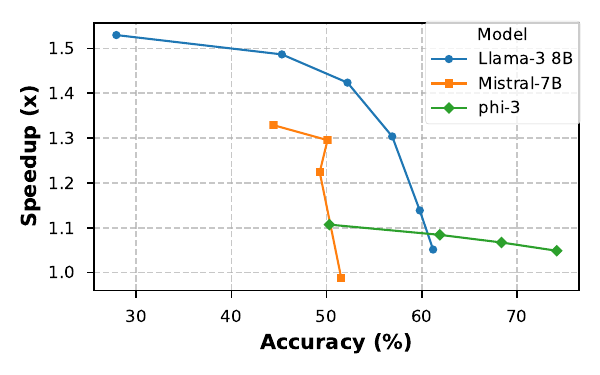}
    \vspace{-10pt}
    \caption{\textbf{Speedup vs. Accuracy}. Different points are from different sparsity percentages.}
    \label{fig:accuracy_latency}
\end{figure}

We use stock PyTorch running in dense format\footnote{PyTorch has a beta API for sparse tensors, but we could not run our model in BF16 or FP16 using it because some operations are not implemented for these formats yet.} as the baseline to ensure a fair comparison and consistency across all other operations, especially since it utilizes AMX when available. Table~\ref{tab:layer_latency} presents the latencies observed for layer 5 of Llama 3 8B when run using stock PyTorch versus our custom sparse kernel. Our kernel outperforms PyTorch across all projections, with performance improvements ranging from 1.22$\times$ in the most time-consuming projection to 2.03$\times$ in the least time-consuming projection. Figure~\ref{fig:accuracy_latency} illustrates the tradeoff between end-to-end speedup and accuracy on GSM8K\cite{cobbe2021training} for multiple models adopted from SQFT\cite{munoz2024sqft}.

\begin{table}[]
\centering
\caption{Speedup in latency of the individual linear modules of layer 5 in Llama 3 8B \newline }
\label{tab:layer_latency}
\begin{tabular}{ccc}
\toprule
Name                  & Dimensions                  & Speedup ($\times$) \\ 
\midrule
q\_proj               & $4096\times4096$             & $1.44$                           \\ 
k\_proj               & $4096\times1024$             & $2.03$                          \\ 
v\_proj               & $4096\times1024$             & $1.41$                         \\ 
o\_proj               & $4096\times4096$             & $1.3$                           \\ 
gate\_proj            & $4096 \times 14336$          & $1.26$                           \\ 
up\_proj              & $4096 \times 14336$          & $1.22$                         \\ 
down\_proj            & $14336 \times 4096$          & $1.36$                           \\ \bottomrule
\end{tabular}
\vspace{-18pt}
\end{table}

\begin{figure*} [t]
    \centering
    \includegraphics[width=1\linewidth]{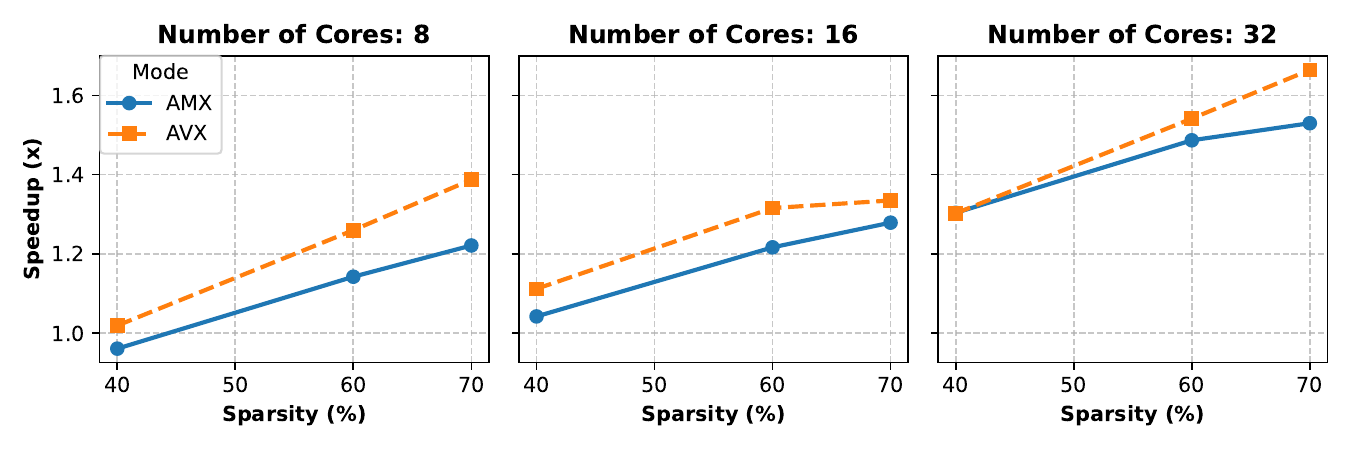}
    \vspace{-20pt}
    \caption{\textbf{Speedup over Stock PyTorch vs. Sparsity}: Llama3 8B decoding with context length 512. The sparse kernels with AVX and AMX have better performance compared to stock PyTorch across the number of cores. As sparsity increases, speedup increases.}
    \vspace{-8pt}
    \label{fig:performance_with_sparsity}
\end{figure*}

For end-to-end speedup, we evaluate the full Llama3 8B model using our AVX and AMX sparse kernels against stock PyTorch. Figure~\ref{fig:performance_with_sparsity} demonstrates how performance scales with sparsity across different numbers of CPU cores. Both AMX and AVX sparse kernels achieve a speedup compared to the stock PyTorch as sparsity increases. The gap between AMX and AVX decreases as the number of cores increases, likely due to reduced cache contention, which decreases memory access bottlenecks for AMX weight construction.

\begin{figure} [t]
    \centering
    \includegraphics[width=.85\linewidth]{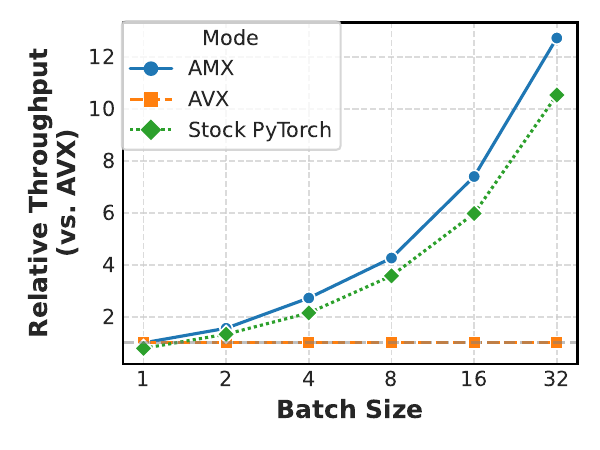}
    \vspace{-10pt}
    \caption{\textbf{Batched Decoding:} AMX is superior to AVX, and our kernel achieves 20.8\% better performance than stock PyTorch at batch size = 32.}
    \label{fig:batched_performance}
    \vspace{-14pt}
\end{figure}

The advantage of AMX becomes apparent at higher batch sizes. Figure~\ref{fig:batched_performance} shows performance across various batch sizes comparing decoding throughput of the stock PyTorch and our AMX kernels to the throughput of our AVX kernel. The Llama3 8B checkpoint was taken from Shears~\cite{munoz2024shears}, which maintains accuracy with 50\% unstructured sparsity. The results indicate that the two AMX kernels achieve higher throughput as batch size increases than the AVX kernel that is designed for \textbf{vector multiplication} rather than \textbf{matrix multiplication} for which AMX is designed.

\begin{figure} [t]
    \centering
    \includegraphics[width=.85\linewidth]{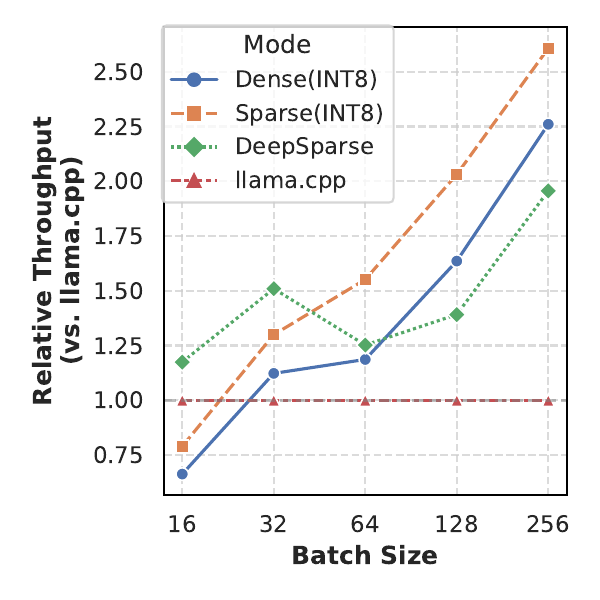}
    \vspace{-10pt}
    \caption{\textbf{INT8 Kernels:} Decoding throughput comparison between our AMX INT8 dense kernel, sparse kernel, DeepSparse and llama.cpp with context length 2 running on 32 CPU cores for Llama 3 7B model with 50\% sparsity. AMX allows us to outperform DeepSparse and llama.cpp at high batch sizes.}
    \label{fig:deepsparse_batched_performance}
\end{figure}

We also compare our kernel to the proprietary DeepSparse engine~\cite{deepsparse} and llama.cpp \cite{llamacpp} in Figure~\ref{fig:deepsparse_batched_performance}. While DeepSparse employs optimizations beyond unstructured sparsity, it is limited to specific models. Our kernel is general-purpose and compatible with any PyTorch model, but the latest Llama model that DeepSparse supports at the time of writing is INT8 Llama 2 7B so we use that model for this comparison. For a fair comparison, we use our INT8 kernel, and to minimize the effect of attention, which is irrelevant, we set the context length to 2\footnote{DeepSparse benchmarking tool does not work out of the box with context length 1.}. The results indicate that our kernel outperforms DeepSparse and llama.cpp at higher batch sizes. This advantage is primarily due to AMX and its ability to perform matrix-matrix multiply rather than vector-vector multiply. Other runtimes like OpenVINO, which also leverage AMX, achieve higher performance but employ additional optimizations, such as operator fusion, which are outside the scope of our framework.

\section{Attention Kernel}
\vspace{-3pt}
In the previous sections, we focused on linear layers. Here, we explore the potential of applying unstructured sparsity to attention computations. During attention, the incoming query value is multiplied by the cached K values, followed by a softmax operation, and the resulting values are multiplied by the cached V values. The cached K and V matrices can be treated as weight matrices and sparsified in an unstructured manner. We first evaluate the impact of unstructured sparsity on accuracy and then adapt the kernel to accelerate the matrix multiplication workload of attention, demonstrating the resulting speedup.

\subsection{Sparsity in the KV Cache}
\vspace{-3pt}

Various methods have been proposed to optimize the KV cache, such as dropping certain tokens~\cite{xiaoefficient, zhang2023h2o}, clustering tokens~\cite{zandieh2024subgen}, or applying channel sparsity~\cite{xu2024think}. In our approach, we apply unstructured sparsity to the KV values using magnitude-based pruning, where values with the lowest magnitudes are dropped within each layer.

\begin{figure} [t]
    \centering
    \includegraphics[width=0.99\linewidth]{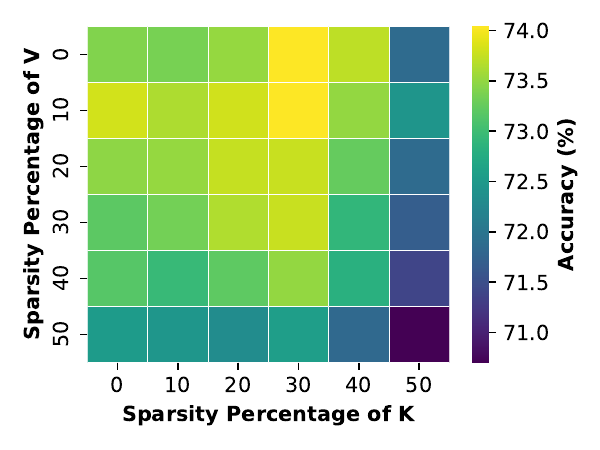}
    \caption{\textbf{KV Sparsity Accuracy} Downstream Accuracy on multiple one-shot tasks for different percentages of KV pruning. Accuracy change is less than 1\% at 30\% K Sparsity and 50\% V Sparsity.}
    \label{fig:oneshot_acc_with_sparsity}
\end{figure}

Figure~\ref{fig:oneshot_acc_with_sparsity} shows the accuracy drop across downstream tasks (calculated as the geometric mean of accuracies for PIQA~\cite{bisk2020piqa}, ARC (Easy \& Challenge)\cite{clark2018think}, BoolQ\cite{clark2019boolq}, HellaSwag~\cite{zellers2019hellaswag}, and WinoGrande~\cite{sakaguchi2021winogrande}). The drop in accuracy is less than 1\% with 30\% K pruning and 50\% V pruning.

\subsection{Acceleration}
\label{sec:acceleration}
\vspace{-3pt}

The sparse kernel introduced in Section~\ref{sec:amx_sparse_kernel} is adapted here for attention computations. Within attention, matrix multiplication is required for computing QK and RV (where R is the result of the scaled softmax of QK). This operation is a batched matrix multiplication with an additional head dimension. The sparse kernel is modified to handle this operation, leveraging the independence of heads to parallelize across them.

To manage the KV cache efficiently, we modify the attention code to initialize an empty cache after prefill, storing all previously cached values in the model state similar to how weights are stored. PyTorch’s native functions for updating the cache and its \textit{repeat\_kv} function (used in Llama 3 8B’s Grouped Query Attention~\cite{ainslie2023gqa}) incur significant overhead with large KV caches due to memory reallocation for each new token. By replacing the cached tokens with our sparse format, which maintains a constant size within the model state, and saving new tokens in a separate dynamic set, decoding becomes over 6$\times$ faster. This enables efficient decoding at high context lengths (e.g., 16K) on CPUs, allowing queries on long contexts with reasonable response times.

\begin{figure} [t]
    \centering
    \includegraphics[width=0.99\linewidth]{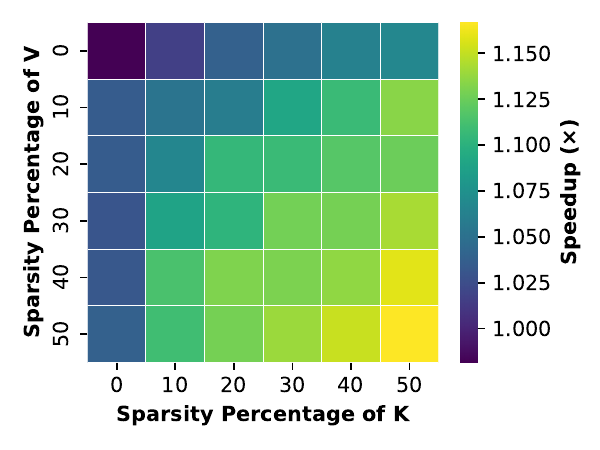}
    \vspace{-10pt}
    \caption{\textbf{KV Sparsity Performance:} End-to-end decode latency speedup when using the sparse kernel for different percentages of KV pruning. Baseline is the dense kernel latency. We are able to achieve 1.14$\times$ speedup with minimal accuracy loss.}
    \label{fig:kv_performance}
    \vspace{-8pt}
\end{figure}

For a fair comparison, we benchmark our sparse kernel against the dense kernel, as shown in Figure~\ref{fig:kv_performance}. Since the dense kernel performs similarly to the stock PyTorch during the decoding stage, this comparison isolates the effect of sparsity. The results show that decode latency improves as the sparsity percentage of K and V increases. At a setting with less than 1\% accuracy loss, we observe a \textbf{1.14$\times$} improvement in latency over the dense kernel.
\vspace{-4pt}

\section{Discussion}

The presented kernels demonstrate that unstructured sparsity can achieve considerable speedup in linear layers. Notably, computations are still performed using dense weights; the speedup is realized by reducing memory transfer in a \textbf{load-as-sparse}, \textbf{compute-as-dense} approach~\cite{xia2023flash}. While this requires additional computation to reconstruct dense weights from the compressed format, the approach is particularly effective in memory-bound scenarios. Conversely, in compute-bound scenarios, applying unstructured sparsity may reduce performance, as confirmed by our results.

We also show that unstructured sparsity can complement other compression techniques, such as quantization. For instance, SQFT~\cite{munoz2024sqft} achieves comparable sparsity levels with INT4 quantization while maintaining accuracy. As shown in Figure~\ref{fig:deepsparse_batched_performance}, our sparse INT8 kernel outperforms its dense counterpart in memory-bound scenarios. However, at higher batch sizes, the dense kernel performs better as the system shifts to a compute-bound regime.

AMX enables efficient inference at high batch sizes, surpassing AVX implementations, including highly optimized proprietary solutions like DeepSparse. However, at lower batch sizes, AMX and AVX performance are similar, with AVX sometimes outperforming AMX. This is because our decompression approach relies on expanding weight metadata into an AVX register, which must then be stored to memory before loading into AMX tile registers due to current architectural limitations. This extra step incurs additional memory operations. Caching mitigates this issue by assuming the weight tile remains in the cache. While our results validate this assumption in most cases, inefficiencies arise potentially because data movement between memory and cache reduces the intended bandwidth savings from compression. Addressing this limitation would require direct data transfer from AVX registers to AMX tiles or instructions to expand bit-masked data directly into AMX tiles.

Given the efficiency of AMX for high-batch inference, Sapphire Rapids CPUs can now feasibly serve certain LLMs without requiring GPUs, potentially reducing cost and power consumption. However, this method requires offline preprocessing to organize weights into the compressed format, including weight metadata, weight values, and thread start indices. Changing the number of threads necessitates recomputation of this representation. Additionally, the attention kernel is not suitable for dynamic KV values but remains effective for cached prompts, such as system prompts or static knowledge bases.

Decoding at long context lengths is highly inefficient in PyTorch, becoming impractical at 16K context length. As discussed in Section~\ref{sec:acceleration}, decoding can be efficiently performed at this length for precomputed cached tokens. For example, companies using CPU-only devices can preload their knowledge base as a cached context in an LLM and direct customer queries to it for faster responses, saving both cost and power.

\section{Limitations}

Our system has several limitations. First, it requires preprocessing time, which, although only a few minutes for 8B models, makes it unsuitable for accelerating dynamically generated sparsity, such as activation sparsity. Additionally, our system currently supports only INT8 and BF16 formats, as AMX units do not natively support more compressed formats like INT4. Extending support to INT4 is feasible by dequantizing INT4 values into INT8 before computation.

While our system supports any PyTorch model and serves as a useful tool for measuring the speedup of unstructured sparsity techniques, other systems, such as OpenVINO, achieve better results by incorporating additional optimizations, such as operation fusion, which are not included in our system. These optimizations make OpenVINO more suitable for production use.

\section{Conclusion}

In this paper, we introduced a system that leverages unstructured sparsity to achieve significant speedup in the decode stage compared to current implementations. Our system is general-purpose, compatible with any PyTorch model, and runs out of the box. It delivers a \textbf{1.42$\times$} performance improvement over the current PyTorch version for Llama 3 8B and, at high batch sizes, achieves over \textbf{1.4$\times$} higher throughput compared to proprietary systems like DeepSparse. Additionally, we demonstrate the potential of using unstructured sparsity in the KV cache to reduce latency per token, achieving a \textbf{1.14$\times$} speedup with simple magnitude pruning.

\clearpage

\bibliography{example_paper}
\bibliographystyle{mlsys2024}

\newpage
\appendix

\section{Sparse Weights Decompression Process}
\label{app:sparse_weights_decompression_process}

We show the steps used to perform weight decompression in Algorithm~\ref{alg:sparse_kernel}.

\begin{algorithm}
\caption{Processing Sparse Format in the Sparse Kernel}
\label{alg:sparse_kernel}
\begin{algorithmic}
\raggedright
\Require 
    $weight\_metadata$: Bitmap indicating non-zero weights
    \newline
    $weight\_values$: Array of non-zero weight values
    \newline
    $weight\_values\_i$: Pointer to the current index in $weight\_values$ for the next non-zero weight
    \newline
    $weight\_tile\_mem$: Memory region for storing weight values
\Ensure 
    $weight\_tile$: AMX tile filled with expanded weights
\Statex
\hspace{-\algorithmicindent} \textbf{Initialization:}  

Fetch 512 bits ($16$ rows $\times$ $32$ bits per row) of $weight\_metadata$ into AVX register $weight\_metadata\_local$.  

Compute prefix sum to get pop count of each element in $weight\_metadata\_local$ into AVX register $weight\_metadata\_local\_popcounts$.  

\Statex
\hspace{-\algorithmicindent} \textbf{Loop through all tile rows:}  
\For{$row\_index = 0$ to $15$}
    \Statex \hspace{\algorithmicindent} \textbf{Step 1: Expand the current row}  
    
    \hspace{\algorithmicindent * 2} Use $vpexpandw$ with $row\_index$ from $weight\_metadata\_local$ and $weight\_values[weight\_values\_i]$ to get 32 weights in dense format in $weight\_tile\_mem$.  
    \Statex \hspace{\algorithmicindent} \textbf{Step 2: Update $weight\_values\_i$}  
    
    \hspace{\algorithmicindent * 2} $weight\_values\_i \gets weight\_values\_i + weight\_metadata\_local\_popcounts[row\_index]$
\EndFor

\Statex
\hspace{-\algorithmicindent} \textbf{Post-Processing:}  

Load data from $weight\_tile\_mem$ into $weight\_tile$.  

\textbf{Return:} $weight\_tile$
\end{algorithmic}
\end{algorithm}

\section{AVX Kernel Optimization}
\label{app:avx_kernel_optimization}

\begin{figure} [t]
    \centering
    \includegraphics[width=1\linewidth]{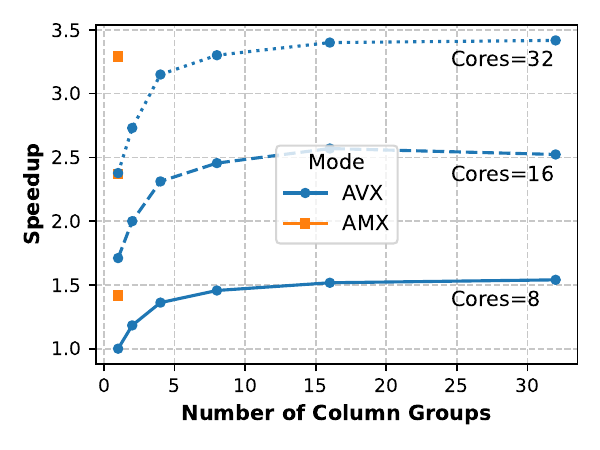}
    \vspace{-25pt}
    \caption{\textbf{Speedup vs. Column Groups:} Speedup for decoding a single token at different number of CPU cores available for AVX implementation having different number of column groups processed together with a single input load. Generally, using more groups leads to better performance, even surpassing that of the AMX implementation. Baseline is 1 column group for AVX kernel on 8 cores.}
    \label{fig:latency_vs_column_groups_vs_amx}
\end{figure}

To enhance utilization, we use multiple AVX registers for both weights and outputs, enabling the input register to be multiplied across all of them. We refer to the number of registers used as $num\_neuron\_groups$. As shown in Figure~\ref{fig:latency_vs_column_groups_vs_amx}, increasing $num\_neuron\_groups$ improves performance, matching or even slightly surpassing the AMX kernel in some cases. This is likely because of the extra memory load and write needed for loading the data into AMX while this step is not needed if the computation happens using AVX registers.

\section{KV Sparsity}

\begin{figure} [t]
    \centering
    \includegraphics[width=1\linewidth]{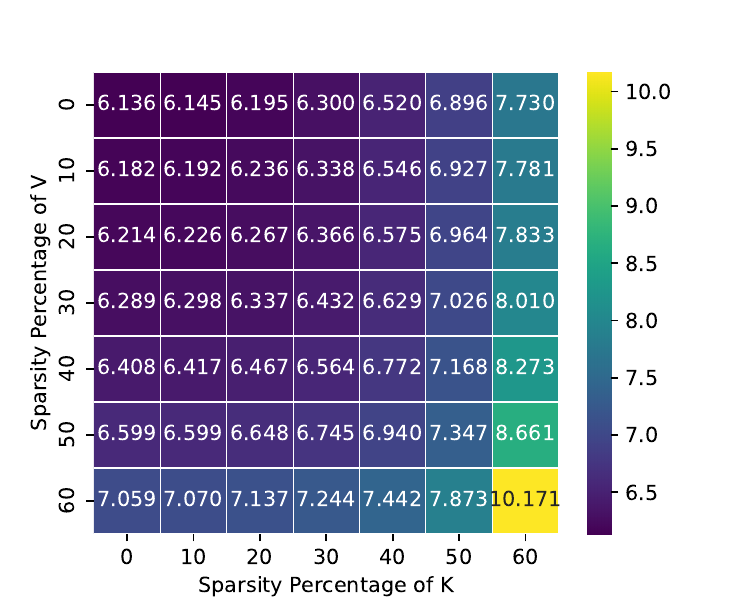}
    \caption{Perplexity on WikiText2 with unstructured sparsity in the KV values.}
    \label{fig:wikitext_ppl_with_sparsity}
\end{figure}

Figure~\ref{fig:wikitext_ppl_with_sparsity} shows how the WikiText2\cite{merity2016pointer} perplexity changes as we apply our suggested unstructured sparsity. We see that at 30\% sparsity in the K and 50\% sparsity in the V, perplexity increases from 6.136 to 6.745.

\begin{figure} [t]
    \centering
    \includegraphics[width=1\linewidth]{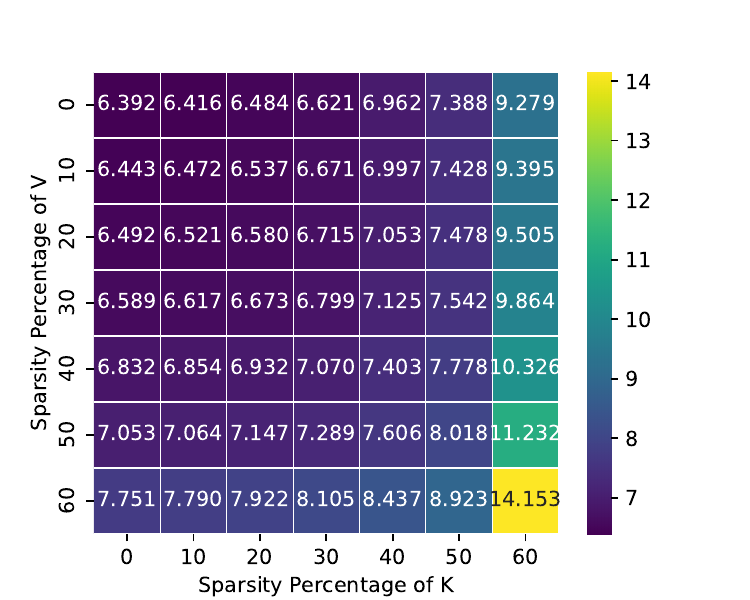}
    \caption{Perplexity on WikiText2 with unstructured sparsity in the KV values after they're quantized to 8 bits.}
    \label{fig:quantization8_wikitext_ppl_with_sparsity}
\end{figure}

Figure~\ref{fig:quantization8_wikitext_ppl_with_sparsity} shows how the perplexity changes for quantized KV. At 30\% K sparsity and 50\% V sparsity, the perplexity difference is still less than 1.

\end{document}